\documentclass[letterpaper, 10 pt, conference]{IEEEconf}
\IEEEoverridecommandlockouts
\usepackage{cite}
\usepackage{amsmath,amssymb,amsfonts}
\usepackage{blindtext}
\usepackage{algorithm}
\usepackage[noend]{algorithmic}
\usepackage{float}
\usepackage{subfig}
\usepackage{graphicx}
\usepackage{textcomp}
\usepackage{xcolor}
\usepackage{authblk}
\usepackage[font=footnotesize]{caption}
\usepackage{xcolor}
\usepackage{url}
\usepackage{changepage}

\def\BibTeX{{\rm B\kern-.05em{\sc i\kern-.025em b}\kern-.08em
    T\kern-.1667em\lower.7ex\hbox{E}\kern-.125emX}}

\newcommand{\real}{\mbox{$I \!\! R$}}
\def \epf{ \hfill $\square$ }
\newcommand{\barr}{ \begin{array}{cl} }
\newcommand{\earr}{ \end{array} }

\begin{document}
\title{\bf\Large Dual Arm Steering of Deformable Linear Objects in 2-D and 3-D \\Environments Using Euler’s Elastica Solutions

\author{A. Levin$^{1}$ and I. Grinberg$^{1}$ and E. D. Rimon$^{1}$ and A. Shapiro$^{2}$}

\thanks{This work receives funding from the European Union's research and innovation program under grant 
no. $101070600$, project SoftEnable. $^1 \!$~Dept. of ME, Technion, Israel. $^2 \!$~Dept. of ME, Ben-Gurion~\mbox{University,}~Israel.}
}

\maketitle

\begin{abstract}
This paper describes a method for steering deformable linear objects using two robot hands in environments populated by sparsely spaced obstacles. The approach involves manipulating an elastic inextensible rod by 
varying the gripping endpoint positions and tangents.
Closed form solutions that describe the flexible linear object shape in~planar~\mbox{environments,} Euler's elastica, are described. The paper uses these solutions~to formulate criteria for non self-intersection, stability and obstacle avoidance. These criteria are formulated as constraints in the flexible 
object six-dimensional configuration space that~represents~the~robot gripping endpoint positions and tangents. In particular,~this~paper introduces a~novel criterion that ensures 
the flexible object 
stability during steering.
All safety criteria are integrated into a~scheme for steering flexible linear objects in planar environments, which is lifted into a~steering scheme 
in three-dimensional environments populated~by~\mbox{sparsely}~spaced~obstacles.  Experiments with a~dual-arm robot demonstrate the method.
\end{abstract}

\vspace{-.055in}

\section*{\textbf{ I. Introduction}}
\vspace{-.055in}

\noindent Robotic manipulation of~\mbox{flexible}~linear~\mbox{objects}~is~\mbox{challenging} due to their multiple degrees of freedom and the need to account for their 
self-intersection and stability. 
Robotic applications include cable routing and untangling 
\cite{cable_routing11,bretl_multiple,cable_rss22}, surgical suturing~\cite{jackson&cavusogl,javdani}, knot tying~\cite{balkcom15,hirai_knot}, compliant mechanisms~\cite{till17}, fresh food 
handling~\cite{food_survey22}
and agricultural robotics~\cite{adaptive_LDO22}. {\small DARPA}'s Plug-Task challenge~\cite{chang&padir}, {\small EU} IntelliMan~\cite{IntelliMan}, {\small EU} SoftEnable~\cite{SoftEnable} and {\small ICRA} workshops~\cite{zhu_survey22,deform_23,deform_24} are all dedicated to robotic handling and manipulation of soft and deformable~objects.

\begin{figure}
\centerline{\includegraphics[scale=0.2]{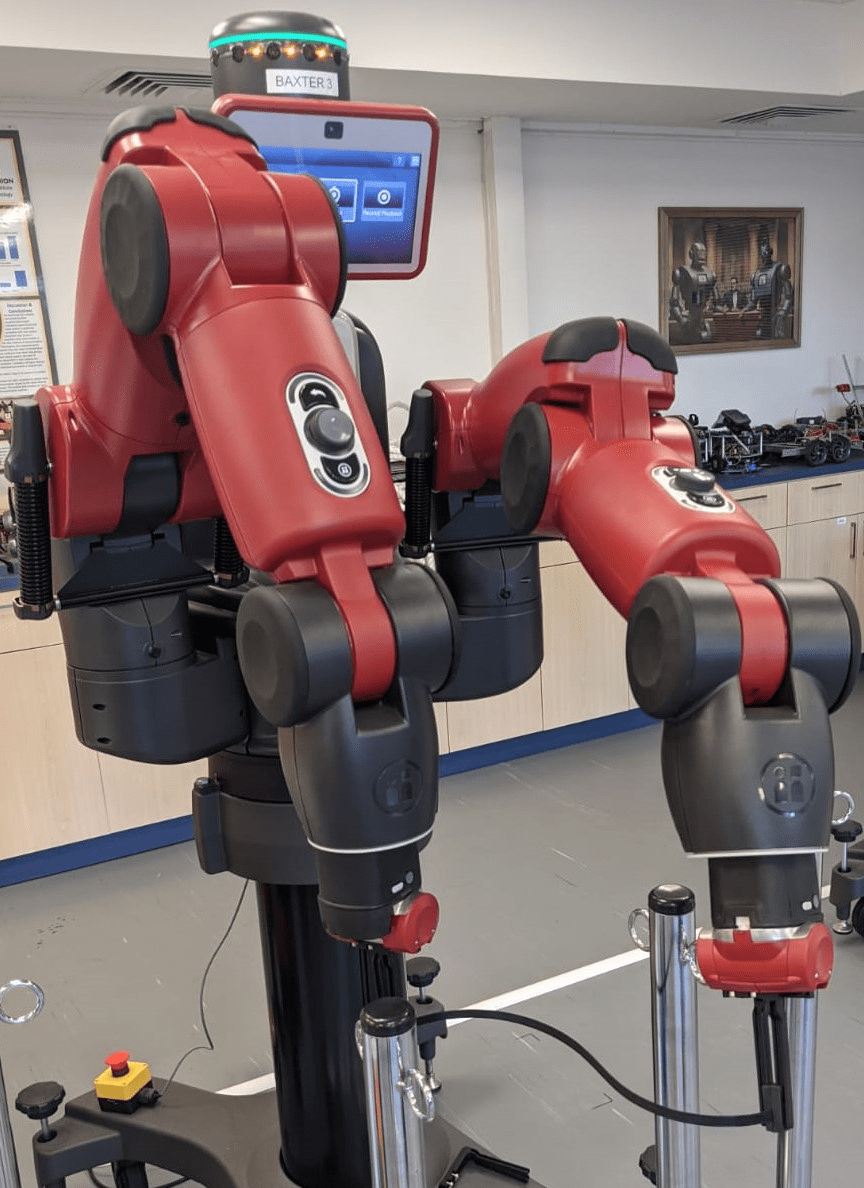}}
\caption{ A dual-arm robot has to steer a~flexible linear object in a~stable and non self-intersecting manner while avoiding sparsely spaced obstacles.}
\label{baxter.fig}
\vspace{-.3in}
\end{figure}

This paper considers robotic steering of flexible linear objects 
in a~stable and non self-intersecting manner while avoiding sparsely spaced obstacles (Fig.~\ref{baxter.fig}). The flexible objects are modeled as inextensible elastic rods having various cross-sections, termed {\em flexible cables.} The paper describes a dual arm steering scheme based on closed-form Euler's elastica solutions in two-dimensions.
The paper then demonstrates how the {\small 2-D} steering scheme can be~lifted~into~\mbox{flexible}~cable steering scheme in three-dimensional environments populated by sparsely spaced obstacles.
The paper assumes the flexible cable elastic energy dominates the cable gravitational energy, thus allowing  lifting the flexible cable manipulation to {\small 3-D} workspace with negligible gravitational effects (Fig. 1).

{\bf Related work:} When a flexible cable is subjected to endpoint forces and moments that maintain its equilibrium state in the plane, its curvature at each point is proportional to the {\em bending moment} at this point according to Euler-Bernoulli bending moment law. 
See Wakamatsu~\cite{wakamatsu04}~for~a~survey of flexible linear object modeling techniques.

In the robotics literature, sampling-based approaches are used to plan flexible cables steering paths. Moll~\cite{kavraki06} samples cable endpoint positions, then computes their stable shapes by numerical optimization of the flexible cable total elastic energy. Bretl~\cite{bretl_ijrr14} uses Sachkov’s theory to describe the flexible cable total elastic energy minimization as an optimal control problem. Bretl shows that the {\em adjoint equation} which describes the flexible cable equilibrium shapes when held by two robot hands is fully determined by a small number of costate variables. In {\small 2-D} environments, these are endpoint forces and moments (three variables), while in {\small 3-D} environments these are endpoint forces and torques (six variables). Bretl~\cite{bretl_ijrr14} used this insight to design sample-based planners that steer flexible cables in 2-D and 3-D settings. However, in order to verify the \textit{stability} of each sampled flexible cable shape, the costate-to-endpoint Jacobian must be computed by solving the {\em conjugate differential equation}. Sintov~\cite{sintov20} built upon this work by introducing a two-stage process. First generating 
stable flexible cable shapes for sampled costates, then determining the 
cable steering path using numerical tests for self-intersection and obstacle avoidance. Yu~\cite{yu23}
models flexible linear objects as spring-mass systems whose
shape is parametrized by $m$ equally spaced features.
A steering path is computed by sampling $\real^{3m}$ and projecting each sample to a~minimum energy shape of the 
spring-mass system. A local {\small MPC} feedback controller then steers the flexible cable 
along the planned path.
The current paper
offers a different approach for steering flexible linear objects using Euler's elastica parameters that fully predict the objects
stable and non-intersecting shapes~in~analytic~closed~form~manner.


\begin{figure}
\vspace{.2in}
\centerline{\includegraphics[width=0.45\textwidth]{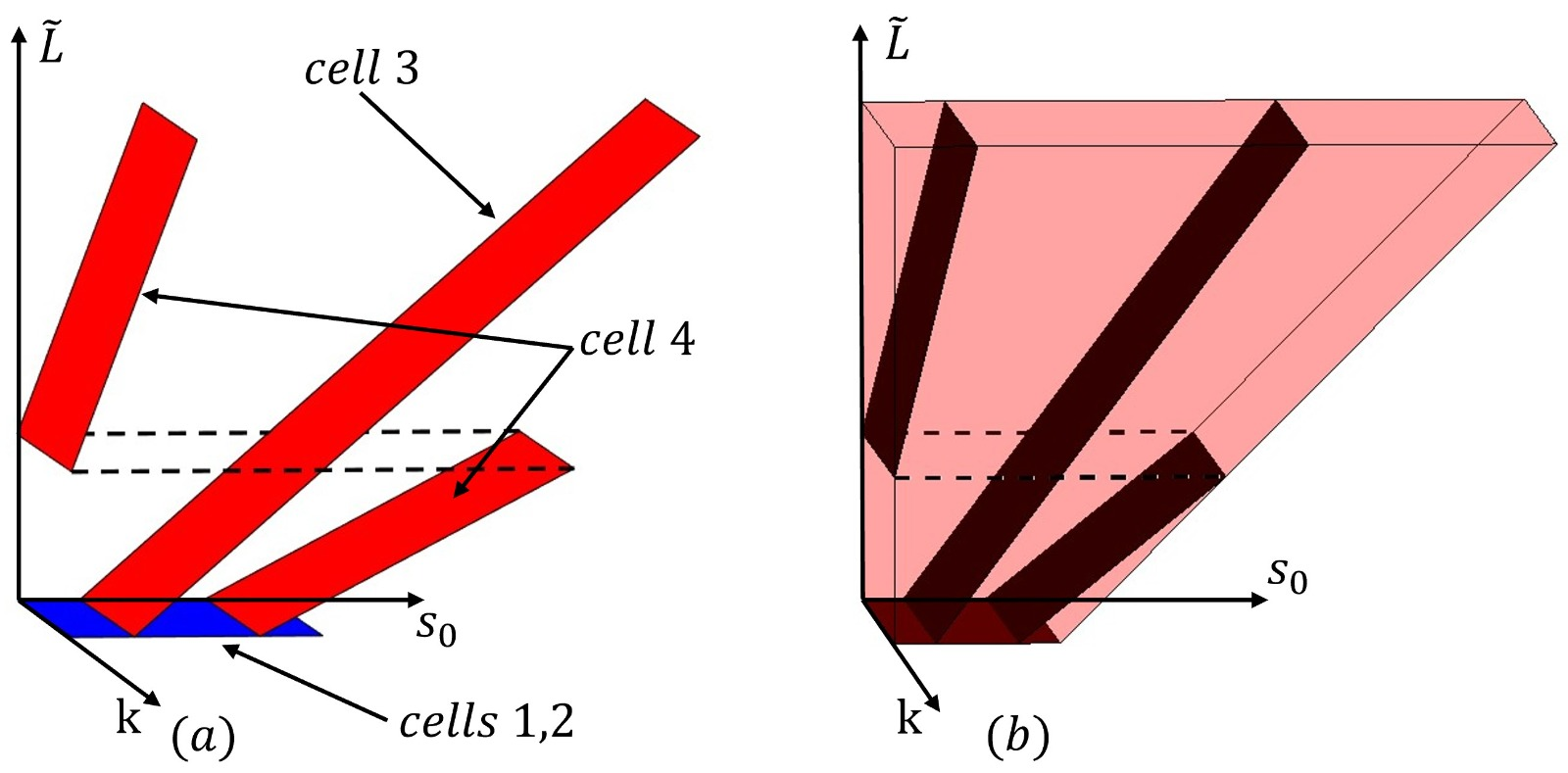}}
\vspace{-.08in}
\caption{The flexible cable elastica parameters describe its equilibrium shapes. (a) Previous work focused on flexible cable held with equal~endpoint~tangen\-ts. (b) This paper considers the full range of elastica parameters where the flexible cable is steered with arbitrary endpoint positions~and~tangents.}
\vspace{-.15in}
\label{elastica.fig}
\vspace{-.18in}
\end{figure}


\textbf{Paper contributions:} This paper steering scheme is based on Euler's elastica solutions for flexible cables equilibrium shapes when held by two robot hands in two-dimensions. The flexible cable configuration space consists of its base-frame position and orientation (three variables), and three elastica shape parameters.  In previous work~\cite{elastica_archive}, we considered flexible cable steering in planar environments using equal endpoint tangents defined by specific planar cells of the flexible cable elastica parameters. These are cells $1,\ldots,4$ in Fig.~\ref{elastica.fig}(a). This paper considers the {\em three-dimensional} elastica parameter space shown in Fig.~\ref{elastica.fig}(b), where the flexible cable is held with arbitrary endpoint positions and tangents during steering. This generalization is achieved by a~novel geometric rule that ensures the flexible cable stability during steering. 

The flexible cable stability and non self-intersection are formulated as constraints in the elastica parameter space. The paper then describes a~piecewise convex representation of the flexible cable equilibrium shapes that is used to efficiently check collision with obstacles during steering. All of these tools are incorporated into flexible cable steering scheme in planar environments.
The paper then demonstrates with experiments how the planar scheme can be used to steer flexible cables in {\small 3-D} environments populated by sparsely spaced obstacles. This scenario is referred to as {\em semi-spatial,} where the flexible cable deformation remains confined to a plane which is free to rotate and translate~in {\small 3-D} space.  The paper thus highlights the availability of {\em a~{\small 9-D} submanifold} (base frame position and orientation and three elastica parameters) embedded in the flexible cable {\small 12-D} configuration space,
that can be used to steer flexible cables in {\small 3-D} space using Euler's elastica solutions.\\
\indent The paper is organized as follows. Section~II~\mbox{summarizes} the flexible cable equilibrium shapes, Euler's elastica, and defines the flexible cable configuration space. Section~III describes a subset of the flexible cable configuration space 
that ensure non self-intersection and stability during steering. Section~IV describes an~efficient collision detection technique, then integrates all tools into 
a~steering scheme.  Section~V describes experiments of the steering scheme in {\small 2-D} and {\small 3-D} obstacle environments. The conclusion suggests future research topics. Two appendices contain proof details
and a~criterion for neglecting 
gravity during {\small 3-D} steering.  

\vspace{-.02in}


\vspace{-.05in}
\section*{\textbf{ II. Flexible Cable Mechanics}}
\vspace{-.02in}

\noindent This section describes Euler's elastica solutions for~flexible cable eq\-uilibrium shapes, the elastica shape parameters, then the configuration space associated~with~these~parameters.

\textbf{Euler’s elastica solutions:} The flexible cable equilibrium shapes are obtained by solving an optimal control problem \cite{bretl_ijrr14}. 
In two-dimensions, we assume an~inextensible flexible cable of length {\small $L$} parameterized by $(x(s),y(s))$ for $s \!\in\! [0, \mbox{\small $L$}]$. The cable's \textit{state variables} are its $(x(s),y(s))$ coordinates and its tangent direction $\phi(s)$, 
forming the state vector $S \!=\! (x,y,\phi)$. The curvature of the cable under arc-length parametrization is $\kappa(s) \!=\! \tfrac{d}{ds}\phi(s)$, which serves as a continuous piecewise smooth {\em control input,} $u(s) \!=\! \kappa(s)$, for the~cable~system equations
\vspace{-.06in}
\begin{equation}  \label{eq.sys}
    \frac{d}{ds}S(s)=
    \begin{pmatrix}
        \Dot{x}(s) \\
        \Dot{y}(s) \\
        \Dot{\phi}(s) 
    \end{pmatrix} =
    \begin{pmatrix}
        \cos\phi(s) \\
        \sin \phi(s) \\
        u(s)
    \end{pmatrix} \hspace{1.5em} \mbox{$s\in [0, L]$.}
\vspace{-.06in}
\end{equation}
\noindent The cable shape, parameterized by arclength in Eq. (1), is determined by its curvature as control input. When the flexible cable modeled as an~elastic rod in a planar environment, its total elastic energy is given by
\vspace{-.08in}
\begin{equation}
    J=\frac{1}{2}EI \!\cdot\!\! \int_{0}^{L} \!\! \kappa^2(s)ds
\vspace{-.06in}
\end{equation}
\noindent where $E>0$ is Young's modulus of elasticity and $I>0$ is the cable cross-sectional second moment of inertia~[26]. The cable stiffness, $EI$, is a known parameter. 

When two robot hands grasp the flexible cable at fixed endpoint positions and tangents, $S(0) \!=\!(x(0),y(0),\phi(0))$ and $S(L) \!=\! (x(L),y(L),\phi(L))$, the flexible cable locally stable shapes form \textit{local minima} of $J$, subject to the endpoint constraints and the cable fixed length constraint described by Eq.~\eqref{eq.sys}. As~$J$ represents bending energy that increases continuously with increasing cable curvature (until reaching plastic yield limit), there always exist stable cable shapes that satisfy the fixed endpoint positions and tangents. The {\em normal Hamiltonian}~\cite{opt_survey} 
(the abnormal case is related to straight line shape~\cite{bretl_ijrr14}) of the cable system defined by Eq.~\eqref{eq.sys} and the elastic energy $J$ is given by
\vspace{-.05in}
\begin{equation}  \label{eq.H}
H(S,\boldsymbol{\lambda},u) \!=\! \lambda_x \!\cdot\! \cos\phi \!+\! \lambda_y \!\cdot\! \sin \phi \!+\! \lambda_\phi u \!+\! \tfrac{1}{2}EI \!\cdot\! u^2
\hspace{2.0em}
\vspace{-.05in}
\end{equation}

\noindent where $\boldsymbol{\lambda}(s) \!=\! ( \lambda_x(s), \lambda_y(s), \lambda_\phi(s))$ are the {\em costate variables.} The costates $\lambda_x,\lambda_y$, and $\lambda_\phi$ correspond to internal force and bending moment along the flexible cable. The equilibrium states of the cable when held by two robot hands are energy extremal cable shapes. The {\em adjoint equation}~\cite{pontryagin} can be used to determine the costate vector $\boldsymbol{\lambda}(s)$ along an~extremal~cable~shape 
\vspace{-.09in}
\begin{equation*}
  \frac{d}{ds}\boldsymbol{\lambda}(s) =-\nabla_S H(S,\boldsymbol{\lambda},u) \hspace{1.85em} \mbox{$S=(x,y,\phi)$}
\vspace{-.06in}
\end{equation*}
\noindent thus leading to the system of adjoint differential equations
\vspace{-.06in}
\begin{equation} \label{eq.adjoint}
\Dot{\lambda}_x \!=\! 0 , \Dot{\lambda}_y \!=\! 0, \Dot{\lambda}_\phi(s) \!=\! \lambda_x(s) \!\cdot\! \sin\phi(s) \!-\! \lambda_y(s) \!\cdot\! \cos\phi(s)
\vspace{-.06in}
\end{equation}
\noindent while the control $u(s) \!=\! \kappa(s)$ satisfies~an~\mbox{additional}~\mbox{condition}
\vspace{-.14in}
\[ 
    \tfrac{\partial H}{\partial u} = 0
\vspace{-.08in}
\]
\noindent thus leading to the algebraic equation
\vspace{-.1in}
\begin{equation} \label{eq.law} 
    \lambda_\phi(s)+ \mbox{\small $EI$}\cdot u(s)=0
\vspace{-.1in}
\end{equation}
\noindent which is Euler-Bernoulli~\mbox{bending}~\mbox{moment}~law.~\mbox{Using}~Eq.~\eqref{eq.law}, $\lambda_x (s)$ and $\lambda_y (s)$~are~\mbox{constant}~along~\mbox{energy}~\mbox{extremal}~\mbox{cable} shapes. These constants define the costate parameters~$\lambda_r$~and~$\phi_0$ 
\vspace{-.06in}
\begin{equation*}
    \begin{pmatrix}
        \lambda_x \\
        \lambda_y
    \end{pmatrix} = \lambda_r \!\cdot\! \begin{pmatrix}
        \cos\phi_0 \\
        \sin\phi_0
    \end{pmatrix}, \ \ \lambda_r = \mbox{\small $\sqrt{\lambda_x ^2+\lambda_y ^2}$} 
\vspace{-.06in}
\end{equation*}

\begin{figure}
\vspace{.1in}
\centerline{\includegraphics[width=0.4\textwidth]{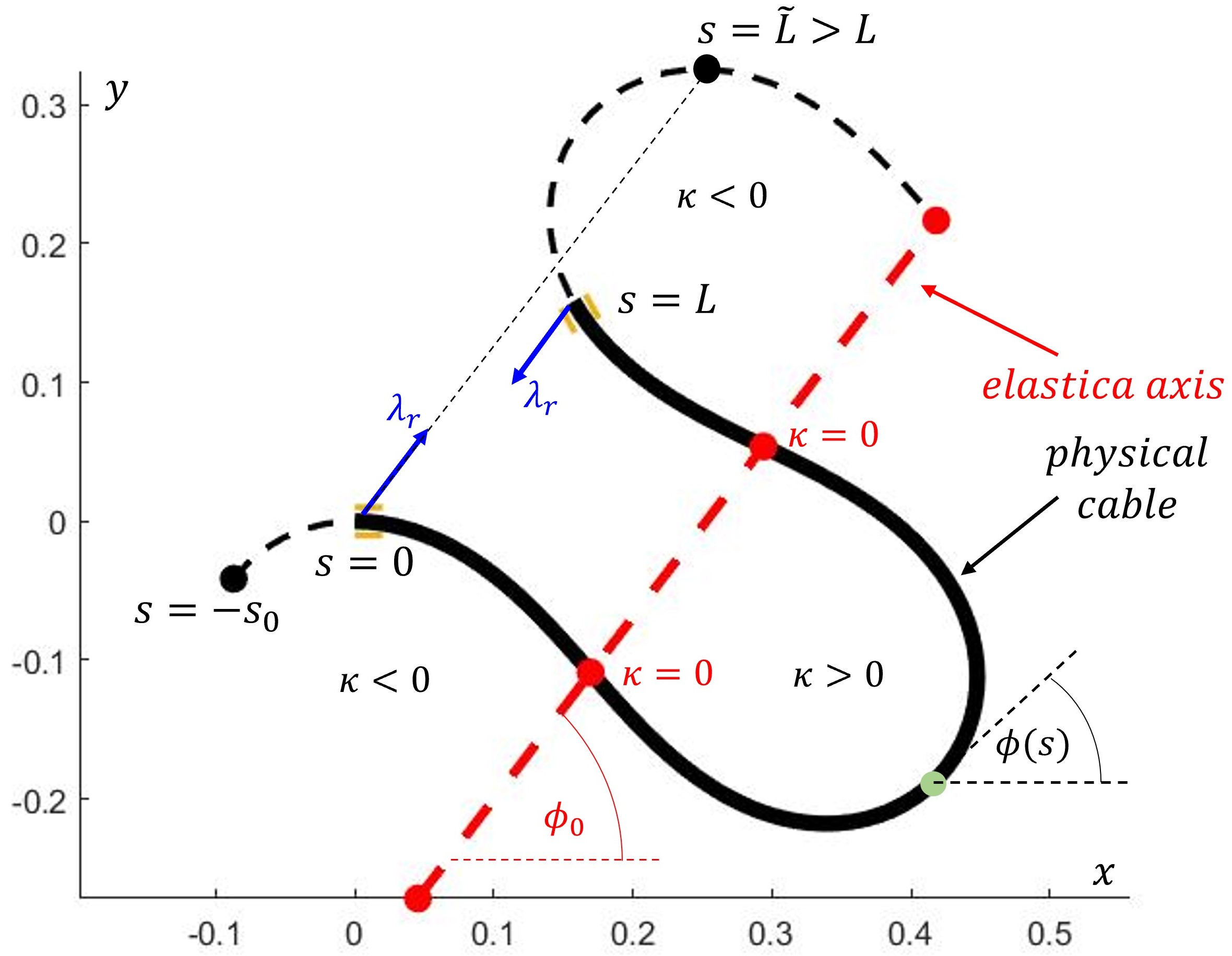}}
\vspace{-.08in}
\caption{Top view of full period elastica shape with the physical flexible cable of length $L$ embedded in its periodic elastica solution of period length~$\tilde{L}$.~The elastica axis with angle $\phi_0$ passes through the zero curvature points~and~is parallel to the opposing forces of magnitude~$\lambda_r$~applied~at~cable~endpoints.}
\label{full_period.fig}
\vspace{-.22in}
\end{figure}

\noindent which represent the magnitude and direction of the force of magnitude~$\lambda_r$ applied at the cable base-frame (blue arrows in Fig.~\ref{full_period.fig}). As the system defined by Eq.~\eqref{eq.sys} is autonomous with no explicit dependence on s, the Hamiltonian remains constant along energy extremal cable shapes, $H(s) \!=\! H^*$ for $s \!\in\! [0, L]$. Substituting $\lambda_\phi (s) \!=\! -EI \!\cdot\! u(s)$ from Eq.~\eqref{eq.law} into $H(s)$ and replacing $u(s)$ by $\kappa(s)$ gives
\vspace{-.11in}
\begin{equation} \label{eq.H*}
    \lambda_r (\cos\phi(s)\cos\phi_0 + \sin\phi(s)\sin\phi_0)-\frac{1}{2}EI \!\cdot\! \kappa^2(s) = H^*
\vspace{-.06in}
\end{equation}
\noindent Taking the derivative of both sides w.r.t. $s$ gives
\vspace{-.1in}
\begin{equation} \label{eq.tderiv}
    \lambda_r (-\sin\phi(s)\cos\phi_0+\cos\phi(s)\sin\phi_0) - EI \!\cdot\! \frac{d}{ds}\kappa(s) = 0
\vspace{-.06in}
\end{equation}
\noindent where we canceled the common factor $\kappa(s) \!=\! \tfrac{d}{ds}\phi(s)$. Substituting the system equations $\dot{x}(s) \!=\! \cos\phi(s)$ and $\dot{y}(s) \!=\! \sin\phi(s)$ into Eqs. \eqref{eq.H*}–\eqref{eq.tderiv} gives
\vspace{-.08in}
\begin{equation*}
    \lambda_r \!\cdot\! R(\phi_0) \!\cdot\!
    \begin{pmatrix}
        \Dot{x}(s)\\
        \Dot{y}(s)
    \end{pmatrix}  = \mbox{\small $ 
    \begin{pmatrix}
        \frac{1}{2}EI\cdot \kappa ^2(s) \!+\! H^* \\
        EI\cdot \frac{d}{ds}\kappa(s)
    \end{pmatrix}    $} 
\vspace{-.06in}
\end{equation*}
where $R(\phi_0) \!\!=\! \mbox{\footnotesize $\begin{bmatrix}
      \cos\phi_0 & \sin\phi_0 \\
      \sin\phi_0 & -\cos\phi_0 
    \end{bmatrix}$}$.
Integrating both sides, $\int_{0}^{s} \Dot{x}(t)dt$ and $\int_{0}^{s} \Dot{y}(t) dt$, gives the flexible cable (x,y) coordinates in terms of its curvature
\vspace{-.02in}
\[ 
    \begin{pmatrix}
        x(s) \\
        y(s)
    \end{pmatrix} \!=\!  \begin{pmatrix}
        x(0) \\
        y(0)
    \end{pmatrix} \!+\! \frac{1}{\lambda_r} R(\phi_0) \!\cdot\!
    \mbox{\small  $\begin{pmatrix}
        \frac{1}{2}EI\cdot\int_0^s \kappa ^2(t)dt+H^*\!  \!\cdot\! s \\[2pt]
        EI\cdot(\kappa(s)-\kappa(0))
    \end{pmatrix}$ }
\vspace{-.04in}
\]
\noindent where $s \!\in\! [0, L]$. This paper focuses on flexible cable~equil\-ibrium shapes that exhibit inflection points, known as {\em inflectional elastica}  (Fig.~\ref{full_period.fig}). 
The curvature of the inflectional elastica is described by an~{\em elliptic cosine function,} $\mathrm{cn}(\cdot,\cdot)$, of the cable path length parameter \cite{love}[p. 402-404] 
\vspace{-.04in}
\begin{equation} \label{eq.curvature}
    \kappa(s)=-2k \sqrt{\lambda} \!\cdot\! \mathrm{cn}(\sqrt{\lambda} \!\cdot\! (s \!+\! s_0), \mathrm{k}) \hspace{1.5em} \mbox{$s\in [0, L]$} 
\vspace{-.06in}
\end{equation}
\noindent where the {\em elliptic modulus parameter,} $0 \! \leq \! \mathrm{k} \! < \! 1$,  is discussed below and $\lambda=\tfrac{\lambda_r}{EI}$. The elliptic cosine function is periodic, similar to the cosine function, and has two zeros per period~\cite{ellip_book}. To get the cable shape, one integrates~Eq.~\eqref{eq.curvature}:
\vspace{-.06in}
\begin{equation} \label{eq.cablexy}
    \begin{pmatrix}
        x(s) \\
        y(s)
    \end{pmatrix} \!=\!  \begin{pmatrix}
        x(0) \\
        y(0)
    \end{pmatrix} \!+\! R(\phi_0) \!\cdot\!
    \begin{pmatrix}
        \Tilde{x}(s) \\
        \Tilde{y}(s)
    \end{pmatrix}
\vspace{-.12in}
\end{equation}
where
\vspace{-.06in}
\begin{equation*}
    \begin{pmatrix}
        \Tilde{x}(s) \\
        \Tilde{y}(s)
    \end{pmatrix} \!=\!  \mbox{\small $ 
    \begin{pmatrix}
        \frac{2}{\sqrt{\lambda}}\big( \epsilon ( \sqrt{\lambda}(s+s_0),\mathrm{k} ) \!-\! \epsilon ( \sqrt{\lambda} \cdot s_0,\mathrm{k} ) \big) -s \\[4pt]
        -\frac{2 \mathrm{k}}{\sqrt{\lambda}} \big(  \mathrm{cn} ( \sqrt{\lambda}(s+s_0),\mathrm{k} ) \!-\! \mathrm{cn} ( \sqrt{\lambda} \cdot s_0,\mathrm{k} ) \big)
    \end{pmatrix} $} 
\vspace{-.06in}
\end{equation*}
where $\epsilon(\cdot,\cdot)=E(\mathrm{am}(\cdot,\cdot),\cdot)$, $\mathrm{am(\cdot,\cdot)}$ is a~{\em Jacobi amplitude function} and $E(\cdot,\cdot)$ is an~{\em incomplete elliptic integral of the second kind}~\cite{ellip_book}. 

\begin{figure}[H]
\centerline{\includegraphics[width=0.44\textwidth]{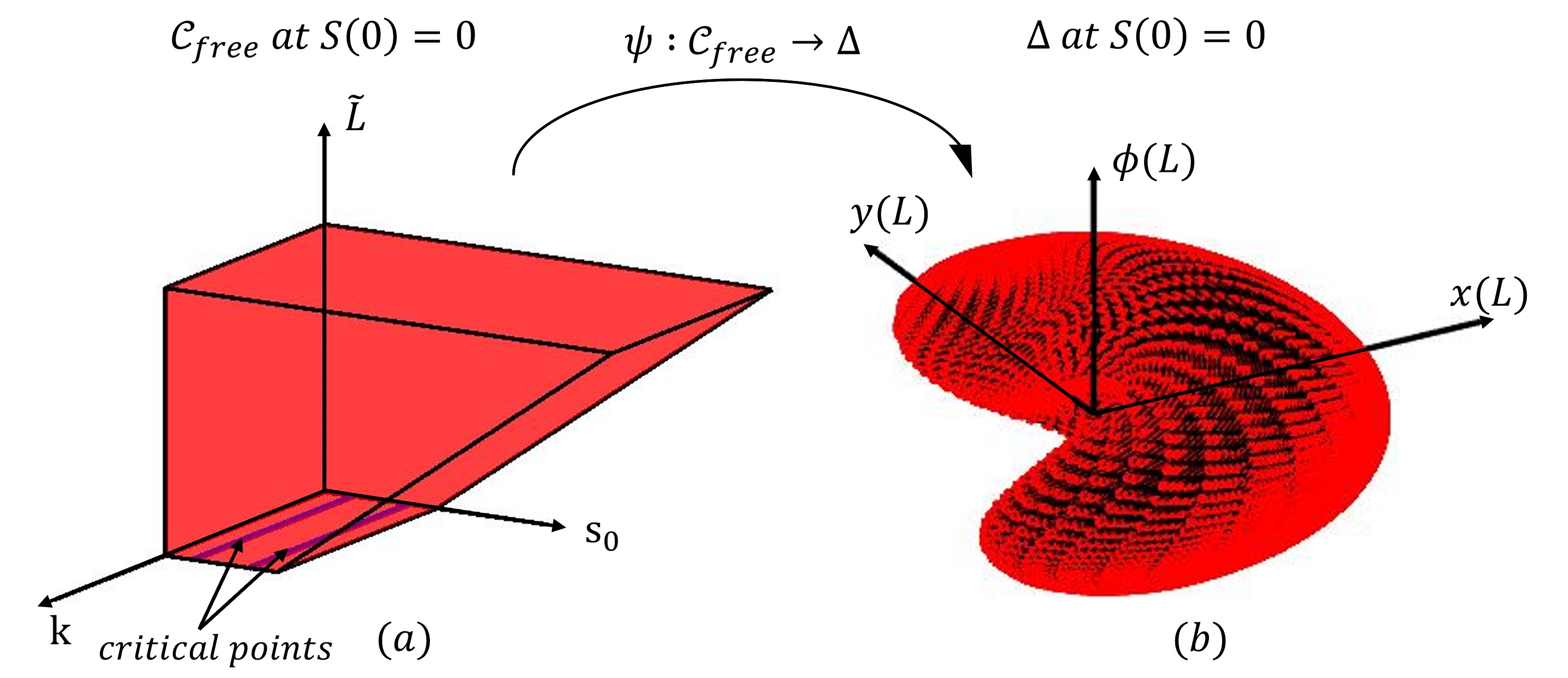}}
\caption{(a) The elastica parameters of $\mathcal{C}_{free}$ when the flexible cable base frame is fixed at the origin, $S(0) \!=\! 0$. (b) Flexible cable distal endpoint, $(x(L),y(L),\phi(L)) \!=\! \psi(q)$ for $q \!\in\! \mathcal{C}_{free}$ with $S(0) \!=\! 0$. Purple lines on bottom of (a) are elastica shapes at which $L \!=\! \Tilde{L}$ while $s_0 \!=\! \tfrac{L}{4}$ or $s_0 \!=\! \tfrac{3L}{4}$. $\mathcal{C}_{free} \subset \mathcal{C}$ is explained in Section~III.}
\label{endpoint_space.fig}
\vspace{-.15in}
\end{figure}

{\bf Flexible cable shape parameters:}
The flexible cable equilibrium shapes can be described by three parameters (Fig.~\ref{full_period.fig}). These are the  modulus parameter, $\mathrm{k}$, a~phase parameter~$s_0$ that measures the physical cable start point and the parameter~$\Tilde{L}$ that measures the full period length of the elastica solution (Fig.~\ref{full_period.fig}). First consider the parameter $\lambda = \tfrac{\lambda_r}{EI}$ in Eq.~\eqref{eq.curvature}. The elliptic cosine function period is $4  K(\mathrm{k})$, where $K(\mathrm{k})$ is the complete elliptic integral of the first kind.\footnote{It is defined as $K(\mathrm{k}) = \int_0^\frac{\pi}{2} \tfrac{d\theta}{\sqrt{1-\mathrm{k}^2 sin^2 \theta}}$, a standard analytic function of the modulus parameter $\mathrm{k}$~\cite{ellip_book}.} 
The argument $\sqrt{\lambda}(s \!+\!s_0 )$ in Eq.~\eqref{eq.curvature} satisfies the full-period relation $\sqrt{\lambda}\cdot \Tilde{L} \!=\! 4 \!\cdot\! K(\mathrm{k})$, which gives
\vspace{-.08in}
\begin{equation}
    \lambda(\mathrm{k},\Tilde{L}) = \bigg{(}\frac{4K(\mathrm{k})}{\Tilde{L}}\bigg{)}^2	
\vspace{-.06in}
\end{equation}
\noindent Next consider the parameter $H^*$ in Eq.~\eqref{eq.H*}. Since $H(s) \!=\! H^*$ for $s \!\in\! [0, L]$, the value of $H^*$ can be~determined~at any point along the cable length. At the zero curvature points, $\kappa(s^*) \!=\! 0$, Eq.~\eqref{eq.H*} gives
$H^*(\mathrm{k},\Tilde{L}) \!=\! \lambda_r \!\cdot\!(\cos\phi(s^*)\cos\phi_0+\sin\phi(s^*)\sin\phi_0 )
 =EI\cdot\lambda(\mathrm{k},\Tilde{L})\sigma(\mathrm{k})$, 
where $\phi_0 \!=\! \phi(0) \!+\! 2 \sin^{-1}(\mathrm{k} \!\cdot\! \mathrm{sn}(\sqrt{\lambda} s_0, \mathrm{k}))$ and $\sigma(\mathrm{k}) \!=\! \cos(\phi(s^*) \!-\!\phi_0)= 1 \!-\! 2 \mathrm{k}^2$, 
Additionally, \cite{elastica_archive} shows that the flexible cable tangent,
$\phi(s)$, is given in terms of
an~{\em elliptic sine function,} $\mathrm{sn}(\cdot,\cdot)$, by
\vspace{-.06in}
\[ 
\phi(s) = \phi_0 - 2\sin^{-1} \left( \mathrm{k} \cdot \mathrm{sn} \mbox{\small $\sqrt{\lambda }$}(s \!+\! s_0), \mathrm{k} \right)  \hspace{.3em} \text{, }  s \in [0, L]
\vspace{-.02in}
\]
\noindent The flexible cable equilibrium shapes are thus determined~by {\em six parameters}: the cable base frame position and orientation, $(x(0),y(0),\phi(0))$, and the elastica parameters $(\mathrm{k},s_0,\Tilde{L})$.
\indent {\bf Flexible cable configuration space:}
The flexible cable shape is fully determined using the base frame position and tangent and the Elastica parameters using Eq.~\eqref{eq.cablexy}.
The flexible cable 
steering path between start and target shapes can be planned in the six-dimensional configuration space defined as $\mathcal{C} \!=\! \{S(0)\times(\mathrm{k},s_0,\Tilde{L}) \in \mathbb{R}^3 \times \mathbb{R}^3\}$, which measures the flexible cable base frame position and orientation, then the modulus, phase and full-period length elastica parameters. The robot {\em feasible gripping states} (endpoint positions and tangents) under inextensible constraint is defined by 
\vspace{-.08in}
\[
\Delta \!=\! \big{\{}(S(0), S(L)) \in \mathbb{R}^3 \times \mathbb{R}^3: \mbox{\small  $\Big{|}\Big{|}
\begin{pmatrix}
x(L) \\ 
y(L)
\end{pmatrix} \!-\!
\begin{pmatrix}
x(0) \\ 
y(0)
\end{pmatrix} \Big{|}\Big{|} $}  \leq L\big{\}}.
\vspace{-.08in}
\]
\noindent The flexible cable steering scheme will use the map $\psi: \mathcal{C} \rightarrow \Delta$, 
which maps configuration space points, $q \!=\!(x(0),y(0),\phi(0),\mathrm{k},s_0,\Tilde{L})$, to feasible gripping states,  $\psi(q) \!=\! (x(0),y(0),\phi(0),x(L),y(L),\phi(L))$, see Fig.~\ref{endpoint_space.fig} for a simplified illustration showing $\Delta$ at $S(0)=0$. \\
\indent In a 3-D workspace, the flexible cable deformation remains planar, while this plane is free to rotate and translate by a 3-D rigid body transformation. Hence, Eq.~\eqref{eq.cablexy} becomes
\vspace{-.08in}
\begin{equation}
    \begin{pmatrix}
        x(s) \\
        y(s) \\
        z(s)
    \end{pmatrix} \!=\!  
    \begin{pmatrix}
        x(0) \\
        y(0) \\
        z(0)
    \end{pmatrix} \!+\! R_x(\phi_x)R_y(\phi_y)R(\phi_0) \!\cdot\! 
    \begin{pmatrix}
        \Tilde{x}(s) \\[2pt]
        \Tilde{y}(s) \\
        0
    \end{pmatrix} 
\end{equation}
where
\begin{equation*}
\vspace{-.08in}
   R(\phi) =\footnotesize{\begin{bmatrix}
        \cos\phi_0 & \sin\phi_0 & 0 \\
        \sin\phi_0 & -\cos\phi_0 & 0 \\
        0 & 0 & 1
    \end{bmatrix}}
\end{equation*}
 $R_x$ and $R_y$ are rotat\-ion matrices around the world frame $x$ and $y$ axes. In the {\small 3-D} case, the flexible cable configuration space is defined by $\mathcal{C}\subset \mathbb{R}^3 \times SO(3) \times \mathbb{R}^3$, and points in $\mathcal{C}$ are $q \!=\! (x(0),y(0),z(0),\phi_x,\phi_y,\phi(0),\mathrm{k},s_0,\Tilde{L})$ where $\phi(0)$ is the tangent direction relative to the flexible cable deformation plane. Appendix~B describes under what conditions the flexible cable elastic energy dominates its gravitational energy.
 \vspace{-.16in}

\vspace{-.06in}
\section*{\textbf{III. The Safe Configuration Space}}
\vspace{-.07in}

\noindent This section characterizes the set of elastica parameters that ensure stability and non self-intersection of the flexible cable  equilibrium shapes in two-dimensions. \\
\indent \textbf{$\!\!$ Stable configuration space:}~Based~on~\cite{gelfand}[theorem 28.1] expressed with optimal control formulation, a necessary and sufficient condition for flexible cable stability is given by three conditions that must be satisfied simultaneously.\\

\vspace{-.15in}
{\indent \textbf{Theorem 1 [local stability]:}
\em{When the following 
conditio\-ns are satisfied, the flexible cable shape forms a~local~minimum of $J$ or equivalently {\bf a~locally stable} equilibrium shape}
\begin{enumerate}
    \item {The triple $(S,\boldsymbol{\lambda},u)$ must satisfy Euler-Lagrange's equation $\Dot{S}^*=\nabla_{\boldsymbol{\lambda}} H^*, \Dot{\boldsymbol{\lambda}}=-\nabla_S H^*$ (energy extremal flexible cable shape).}
    \item {Along the extremum cable shape, $\tfrac{\partial H}{\partial u} = 0$ and $\tfrac{\partial ^2 H}{\partial u^2} \!>\! 0$ 
    (Legendr's necessary condition for a local minimum).}
    \item {The flexible cable shape contains \textit{no conjugate point} in the interval $[0, L]$.}
\end{enumerate}}
 \noindent where $H(S, \boldsymbol{\lambda}, u)$ is the flexible cable Hamiltonian defined by Eq.~\eqref{eq.H}.

\begin{figure}
\vspace{.2in}
\centerline{\includegraphics[width=0.42\textwidth]{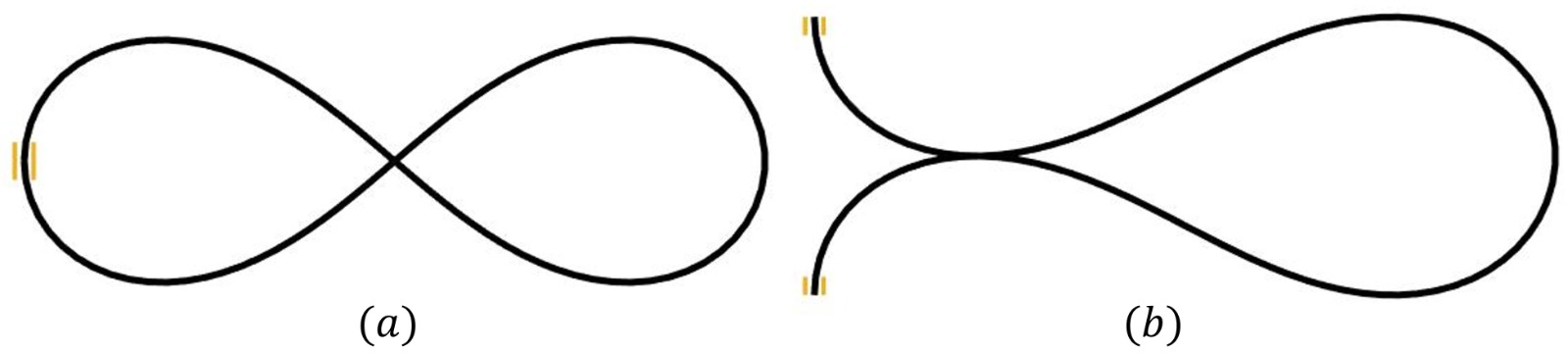}}
\vspace{-.1in}
\caption{(a) The limit of stable full-period cable shapes occurs at $\mathrm{k}_c \!=\! 0.908$, which forms a~figure eight. (b) The limit of non self-intersecting cable shapes occurs at $\mathrm{k}_{max} \!=\! 0.855$, at which the full-period elastica just touches itself.}
\label{crit_k.fig}
\vspace{-.25in}
\end{figure}

The next lemma asserts the stability of {\em full period} elastica shape. The lemma uses the critical modulus parameter, $\mathrm{k}_c \!=\! 0.908$, at which the elastica full-period forms~a~figure eight shape (Fig.~\ref{crit_k.fig}(a)). \\
\indent  \textbf{Lemma 1:} 
{\em{All full period elastica shapes with $L \!=\! \Tilde{L}$ and $\mathrm{k} \!\in\! [0,\mathrm{k}_c]$ are locally stable, except full period shapes having phase parameter $s_0 \!=\! \frac{L}{4}$ or $ \frac{3L}{4}$}}.\\
The proof of Lemma 1 is found in Appendix 1. 
Lemma 1 is now used to establish the stability of all sub full-period cable shapes with $L \!<\! \Tilde{L}$ such that $\mathrm{k} \!\in\! [0,\mathrm{k}_c]$. \\
\indent \textbf{Theorem 2 [locally stable cable shapes]:} {\em Using flexible cable configuration parameters, $(x(0),y(0),\phi(0),\mathrm{k},s_0,\mbox{\small $\Tilde{L}$}) \in \mathbb{R}^6$, the subset of the cable equilibrium shapes given by
\vspace{-.11in}
 \begin{multline} \label{eq.Cstable}
    \mathcal{C}_{stable} = \big\{(x(0),y(0),\phi(0),\mathrm{k},s_0,\Tilde{L})\in \mathbb{R}^6:0\leq \mathrm{k} \!<\! \mathrm{k}_c \\[-2pt],0\leq s_0  \!<\! \Tilde{L},
    \Tilde{L}\geq L,\ at\ \Tilde{L}=L, s_0\neq \tfrac{L}{4} \ or \ \tfrac{3L}{4}\big\}
\end{multline}

\vspace{-.09in}
\noindent form locally stable Elastica shapes.} 

\indent {\bf Proof:}
Consider the three conditions of Theorem~1.  One can verify by direct substitution that Euler's elastica  form
non-trivial solutions $(S, \boldsymbol{\lambda})$ of the adjoint equations~\eqref{eq.adjoint}, as well as solutions of Eqs.~\eqref{eq.law}  and~\eqref{eq.curvature} 
where $u(s) \!=\! \kappa(s)$ for $s \!\in\! [0, \mbox{\small $L$}]$. Hence, according to the first item in Theorem~1, the solution $(S, \boldsymbol{\lambda})$ is an~extremum.  Next, in the case of flexible cables, $\tfrac{\partial ^2 H}{\partial u ^2} \!=\! \mbox{\small $EI$} \!>\! 0$, as required by Legendre's necessary condition.
Let us now focus on the third item of Theorem~1, showing that all flexible cable shapes with parameters specified by Eq.~\eqref{eq.Cstable} have {\em no conjugate points}.  

Consider the local stability of {\em full-period} elastica shapes, with length $L \!=\! \Tilde{L}$ such that
$\mathrm{k} \!\in\! [0,\mathrm{k}_c]$ (Lemma~1).
The full-period elastica parameters form {\em a~subset} of the elastica parameters specified by Eq.~\eqref{eq.Cstable}. This subset, $\mathcal{C}_{full-period} \subset  \mathcal{C}_{stable}$,  consists of elastica parameters $\mathrm{k} \!\in\! [0,\mathrm{k}_c]$ and  $s_0 \!\in\! [0, L]$ such that $L \!=\! \Tilde{L}$ (physical cable length equals elastica full period length). 
All other elastica shapes in $\mathcal{C}_{stable}$ specified by Eq.~\eqref{eq.Cstable}
form {\em sub full-period shapes,} with  $L \!<\! \Tilde{L}$. When the physical cable is {\em shorter} than the elastica full period length, $L \!<\! \Tilde{L}$, the {\em principle of optimality}~\cite{liberzon}[Section 5.1.2] 
also holds for the flexible cable system of Eq.~(1). Based on this principle, when the full-period 
elastica shape that starts at $s \!=\! s_0$ and ends at $s \!=\! s_0 \!+\! \Tilde{L}$ is {\em optimal,} every sub full-period shape that starts at $s \!=\! s_0$ and ends at
$s \!=\! s_0 \!+\! L$ such that $L \!<\! \Tilde{L}$
is also optimal. 
Hence any sub full-period shape occupied by the physical cable, $s_0 \!\leq\! s \! \leq \! s_0 \!+\! L$, contains {\em no conjugate point} to its start point $s \!=\! s_0$. Thus, all three conditions of Theorem~1 are satisfied.~\epf 


\textbf{Remark:} Theorem~2 is consistent with Sachkov \cite{sachkov_conjugate}[Corollary 3.3]
and captures local stability 
of the {\em subset} of flexible cable shapes used in  this paper steering scheme.~\hfill $\circ$ 

\vspace{.02in}
\textbf{Non self-intersection configuration space:}
The non self-intersection condition depends solely on the modulus parameter $\mathrm{k}$~\cite{elastica_archive}. Consider the interval  $\mathrm{k} \!\in\! [0,\mathrm{k}_{max}]$, where at $\mathrm{k}_{max} \!=\! 0.855$ the flexible cable touches itself for the {\em first time} along the elastica full period shape (Fig.~\ref{crit_k.fig}(b)). 
When $0 \!\leq\! \mathrm{k} \!\leq\! \mathrm{k}_{max}$, the flexible cable is locally stable and has no self-intersection. This conservative subset of the flexible cable c-space,  $\mathcal{C}_{free} \!\subset\!\mathcal{C}_{stable}$, is given by 
\vspace{-.06in}
\[ 
\barr
\mathcal{C}_{free} = &
\big\{ \mbox{\small 
 $(x(0),y(0),\phi(0),\mathrm{k},s_0,\Tilde{L})$} \in \mathbb{R}^6:0\leq \mathrm{k} \!<\! \mathrm{k}_{max},\\[2pt]
 &   0\leq s_0  \!<\! \Tilde{L},
    \Tilde{L}\geq L,\ \mbox{at}\ \Tilde{L}=L, s_0\neq \tfrac{L}{4} \ \mbox{or} \ \tfrac{3L}{4}  \big\}
\earr 
\]
\vspace{-.1in}

\noindent When the flexible cable path is planned in $C_{free}$, it is kept stable and non self-intersecting as next discussed.

\vspace{-.02in}
\section*{\textbf{IV.~Cable~Steering~in~Presence~of~Obstacles}}
\vspace{-.04in}

\noindent This section describes an efficient collision detection technique for the flexible cable when steered  among polygonal obstacles, then describes a scheme that plans the flexible cable steering path in the presence of 
obstacles. 
\vspace{.03in}

        \begin{algorithm} 
        \textbf{Input:} vector $q \in \mathcal{C}_{free},$ cable length $L$
        \caption{Bounding Triangles Computation}\label{alg:cap} 
            \begin{algorithmic}[1]
                \STATE Initialize($v_0 = (q_1, q_2)$, $\phi_0=q_3$, $V=v_0$, $s^*=\infty$)
                \IF{$q_5 < \frac{q_6}{4}$ and $q_5+L > \frac{q_6}{4}$} 
                    \STATE $s^*\gets \frac{q_6}{4}$
                \ELSIF{$q_5 < \frac{q_6}{2}$ and $q_5+L > \frac{q_6}{2}$} 
                    \STATE $s^*\gets \frac{q_6}{2}$
                \ELSIF{$q_5 <  \frac{3 q_6}{4}$ and $q_5+L >  \frac{3 q_6}{4}$} 
                    \STATE $s^*\gets \frac{3q_6}{4}$
                \ELSIF{$q_5 <  q_6$ and $q_5+L >  q_6$} 
                    \STATE $s^*\gets q_6$
                \ENDIF
                \WHILE{$q_5 <  s^* <  q_5+L$}
                   \STATE $v_{i+1}\gets (x(s^*-q_5),y(s^*-q_5))$  
                    \STATE $\phi_{i+1}\gets \phi(s^*-q_5)$ 
                    \STATE $m_i \gets line(v_i, \phi_i) \cap line(v_{i+1}, \phi_{i+1})$
                    \STATE $\mathcal{V} \leftarrow V\cup \{m_i,v_{i+1}\}$
                    \STATE $s^* \gets s^*+\frac{q_6}{4}$  
                    \STATE $i \gets i+1$ \\
                    \ENDWHILE
                \STATE $n \gets i,v_n \gets (x(L),y(L)),\phi_n = \phi (L)$
                \STATE $m_{n-1} \gets line(v_{n-1},\phi_{n-1}) \cap line(v_n, \phi_n)$
                \STATE $\mathcal{V} \gets \mathcal{V} \cup {m_{n-1},v_n}$
                \RETURN $\mathcal{V}$ \\
               $\!\!\!\!$  \mbox{\hspace{-1.5em}} \textbf{Output:} triangle vertices $\mathcal{V} \!=\! \{v_0,m_0,\ldots,m_n,v_n \}$, $n \!\leq\! 5$.
            \end{algorithmic}
        \end{algorithm} 

\vspace{-.04in}

\begin{figure} 
\vspace{.15in}
\centerline{\includegraphics[width=0.32\textwidth]{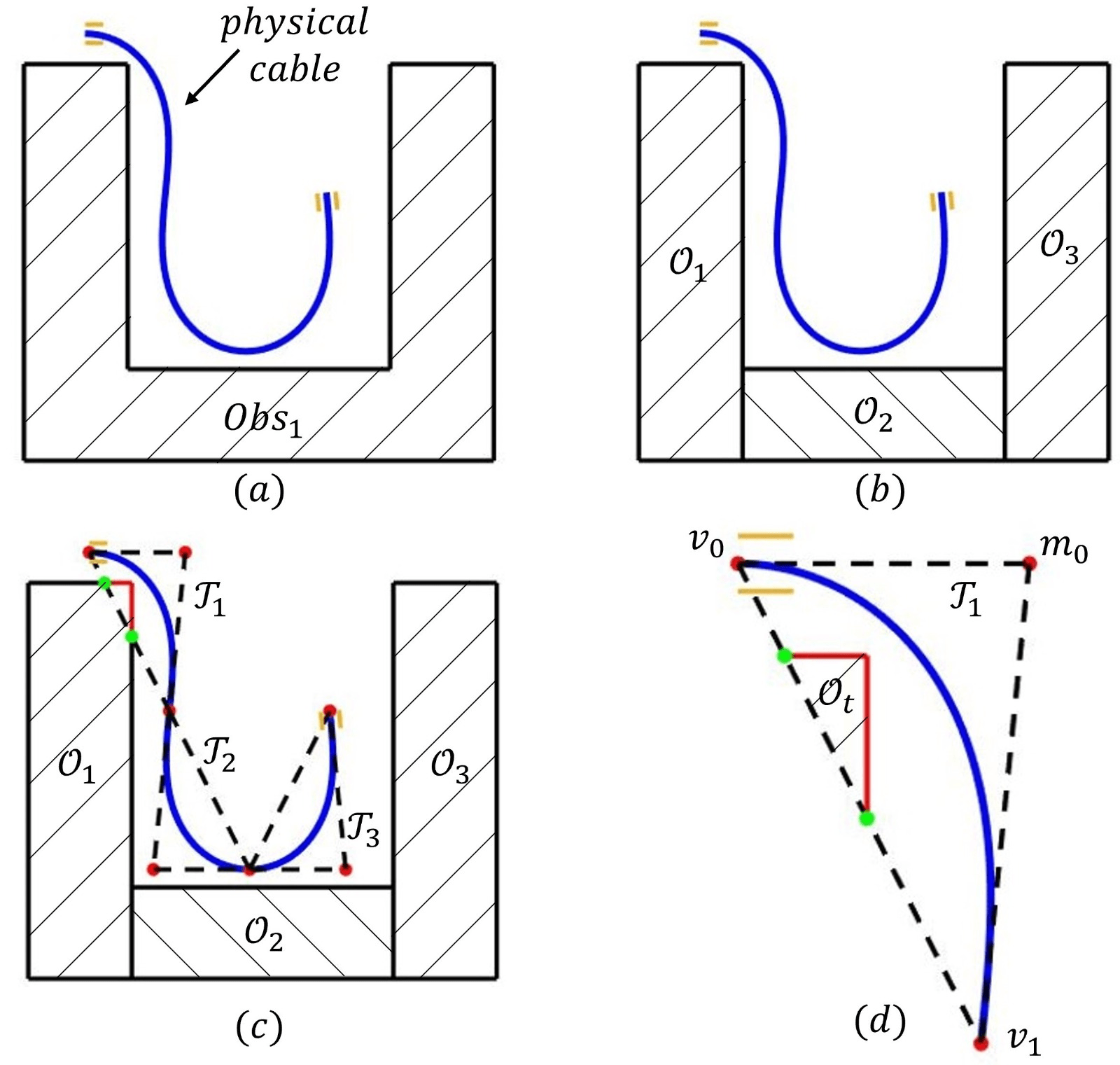}}
\vspace{-.08in}
\caption{Collision checker: (a) Physical cable in blue and obstacle in black. (b) Split obstacle into convex pieces $O_1, O_2, O_3$. (c) Partition flexible cable into convex arcs then bound arcs by triangles $\mathcal{T}_1,\mathcal{T}_2,\mathcal{T}_3$. Green points are intersection points between $\mathcal{T}_i$ and $O_j$ pairs, red lines are $O_t \!=\! (O_j\cap \mathcal{T}_i)$.~(d)~The arc inside  $\mathcal{T}_1$ 
is used to check intersection with the convex obstacle piece~$O_t$.}
\label{collision.fig}
\vspace{-.15in}
\end{figure}

{\bf Obstacles collision checker:}
The collision checker operates in four stages
illustrated in Fig.~\ref{collision.fig}. Stage~I splits the polygonal obstacles into convex pieces ($O_1,O_2,O_3$ in Fig.~\ref{collision.fig}(b)). This stage is carried out only once.~In~Stage~II, for a~given flexible cable shape determined by specific elastica parameter values, the elastica solution is used to split the cable into at most five {\em convex arcs,} then bound each arc by a~triangular region ($\mathcal{T}_1,\mathcal{T}_2,\mathcal{T}_3$ in Fig.~\ref{collision.fig}(c)). 
The first and third vertices of each triangle, $v_i$ and $v_{i+1}$, are either one of the cable endpoints, extremum curvature points, or inflection points along the flexible cable shape, all known in closed form using the elastica solutions at the base frame $S(0)$. The middle vertex of each triangle, $m_i$, is located at the intersection of the tangent lines to the flexible cable arc at $v_i$ and $v_{i+1}$ (Fig.~\ref{collision.fig}(b)). The flexible cable bounding triangles computation is summarized as  
Algorithm~1.


\vspace{.08in}

\begin{algorithm}
\textbf{Input:} the vector $q\in \mathcal{C}_{free}$, cable length $L$, convex obstacle pieces $Obs=\{O_1,O_2,\ldots,O_m\}$.
\caption{Obstacle Collision Checker}\label{alg:cap}
    \begin{algorithmic}[1]
        \STATE $check \gets FALSE$ 
        \STATE $\{ \mathcal{T}_1,…,\mathcal{T}_n \} \gets$ GetBoundingTriangle$(q,L)$ 
        \FOR{$i = 1 \ \mbox{to} \ n$}
        \FOR{$j = 1 \ \mbox{to} \ m$}
            \STATE $O_t = O_j\cap int(\mathcal{T}_i)$ 
            \IF{$O_t \neq \emptyset$}
                 \IF{$O_t \cap ConvArc(t_i) \neq \emptyset$}
                    \RETURN $check \gets TRUE$
        \ENDIF
        \ENDIF
        \ENDFOR
        \ENDFOR 
    \end{algorithmic}
\end{algorithm} 

Stages~III and~IV use the flexible cable bounding triangles as follows. Stage~III checks~intersection~of~the~flexible~cable bounding triangles against each convex obstacle piece. Stage~IV computes the portion of each convex obstacle piece that lies inside the flexible cable bounding triangles (red portion of $O_1$ in Fig.~\ref{collision.fig}(d)).
Then tests for intersection using minimum distance between the cable convex arc inside the triangle and the obstacle boundary in this triangle.~The~collision checker 
is summarized as Algorithm~2.
{\bf Flexible cable steering scheme:}
The flexible cable steering scheme implements a \textit{ weighted $\mbox{\small $A$}^*$} algorithm in the flexible cable
configuration space.
The scheme computes a~steering path in 
a~{\small 6-D} configuration space for planar steering and a~{\small 9-D} configuration space for semi-spatial steering. 
To save storage space, our $\mbox{\small $A$}^*$ 
implementation constructs only a~graph embedded in the flexible cable configuration space during steering path computation. For each grid cell $q\in \mathcal{C}_{free}$, its $12$ 
or $18$ neighbors describe local increase or decrease in c-space coordinates. Using $q \!=\!(x(0),y(0),\phi(0),\mathrm{k},s_0,\Tilde{L})$ the cost function of a node $i$ used by the weighted $\mbox{\small $A$}^*$~is~given~by
\vspace{-.16in}
\[ 
 \mathrm{Cost}(q_i)=(1 \!-\! w)\cdot \mbox{\small $G$}(q_i,q_{i-1})+w\cdot h(q_i,q_T) 
 \vspace{-.07in}
\]
\noindent where $\mbox{\small $G$}(q_i,q_{i-1}) \!=\! ||\psi(q_i) \!-\! \psi(q_{i-1})|| \!+\! \mbox{\small $G$}(q_{i-1})$ such that $||\psi(q_i) \!-\! \psi(q_{i-1})||$ is Euclidean norm,
while $w \!\in\! [0,1]$ is a~tuning parameter.
The 
function $h(q_i,q_T)$ estimates the distance from $q_i$ to the {\em target} cable configuration, $q_T$, measured in terms of minimum robot hands movement by the norm
\vspace{-.08in}
\[ 
h(q_i,q_T) \!=\! ||\psi(q_i) \!-\! \psi(q_T) ||
\vspace{-.06in}
\]
\noindent where $\psi(q)$ has been defined in Section~III.
The tuning parameter in Eq. (17), $w \!\in\![0,1]$ affects the {\em greediness} of the $A^*$ search algorithm. When $w \!=\! 1$, the $\mbox{\small $A$}^*$ algorithm becomes highly greedy with reduced execution times.
When $w \!=\! 0$, the $\mbox{\small $A$}^*$ algorithm becomes Dijkstra's algorithm that opens all neighbors of the current node. In real-world applications execution time is more important than optimality, hence $w \!=\! 0.88$ is used in our implementation.
The flexible cable steering scheme is summarized as Algorithm~3. The algorithm accepts as input start and target cable shapes $q_S,q_T \in \mathcal{C}_{free}$ and computes the flexible cable path in the presence of obstacles.

    \begin{algorithm}
       \textbf{Input:} start and target $S,T \in \mathcal{C}_{free}$, obstacles $O_1,\ldots,O_N$.
        \caption{Flexible Cable Steering Scheme}\label{alg:cap}
        \begin{algorithmic}[1] 
            \STATE Set $open=[q_s], close=[\emptyset] $ split obstacles to convex pieces $\{O_1,O_2,\ldots,O_n \}$  
            \WHILE{$open \neq \emptyset$}
               \STATE $q^* \gets Min(Cost(open))$ 
               \IF{$q^* = T$}
               \RETURN $\mathcal{P}_{best} = backpointer(q^*)$ \COMMENT{path from S to T}
               \ENDIF
               \STATE $close \gets q^*$
               \FOR{$q_i \in \mathcal{C}_{free} $ and\ $ q_i \notin close$ \ and \ $q_i \in neighbor(q^*)$ }
                    \IF{$q_i \notin open$}
                        \IF{ObstacleCollision($q_i,ConvObs, L$)=FULSE}
                            \STATE $open \gets q_i$
                        \ELSE 
                            \STATE $close \gets q_i$
                        \ENDIF
                    \ELSIF{$Cost(q_i) < Cost(open(i))$} 
                        \STATE $open(i) = q_i$
                    \ENDIF
                \ENDFOR   
            \ENDWHILE 
            \RETURN $\mathcal{P}_{best} = \emptyset$ \COMMENT{if no path exist} \\
     \mbox{$\!\!\!\!$ \hspace{-2em} }   \textbf{Output:} path $\mathcal{P}_{best}$ from $S$ to $T$.
            \end{algorithmic}
    \end{algorithm} 


\vspace{-.04in}
\section*{\textbf{V. Representative Experiments}}
\vspace{-.02in}

\noindent To demonstrate the flexible cable steering scheme, we performed two proof-of-concept experiments using the dual arm Baxter robot. 
The flexible object used in the experiments is
a~plastic zip-tie of length $\mbox{\small $L$} \!=\! 80$~$\mathrm{cm}$. 
The steering scheme was implemented using {\small MATLAB} R2023a with software library Elfun18~\cite{batista_matlab} to compute elliptic functions and elliptic integrals.
The code ran on Dell OptiPlex 7000 with 16GB RAM and Intel core I7-12700 2.1~GHz CPU. 
The steering scheme received as input the base frame position and elastica parameters of the zip-tie start and target positions. The output is the trajectory of the robot arms joint angles.  \\
\indent In the steering experiments, the range of the elastica parameters $\mathrm{k}$ and $s_0$ was defined based on the configuration space $\mathcal{C}_{free}$. The elastica parameter $\tilde{L}$ was constrained to the range $[L, 4L]$, allowing the flexible zip-tie shapes to vary between full and quarter period of the full elastica period. The range of the base frame position and orientation was determined by the Baxter robot workspace. 
The resolution of the configuration parameters
was: $\Delta x(0) = \Delta y(0)=\Delta z(0)=0.01$~$\mathrm{m}$, $\Delta \phi(0)=\Delta \phi_x=\Delta \phi_y=2^{\circ}$, $\Delta \Tilde{L} = 0.03L$, $\Delta s_0 = 0.01L$, $\Delta \sigma=0.015$, where the parameter $\sigma$ is defined as $\sigma \!=\! 1 \!-\! 2\mathrm{k}^2$. \\
\indent The Baxter robot has seven joints per arm. One difficulty during the experiments was the lack of a~readily available inverse kinematics (IK) solver for this robot. In order to analytically solve the robot IK problem, we fixed its third joint in each arm which significantly reduced the robot workspace. During path planning, a check was made to ensure that the zip-tie endpoints lie in the robot workspace~\cite{baxter} by checking for the existence of feasible {\small IK} solutions.
The robot arms occupy large volumes, hence many {\small IK} solutions were excluded in order to prevent inter-arm collision.\\
\indent The steering scheme proof-of-concept consists of two experiments. First steering the flexible zip-tie in a~planar obstacle environment (Fig. 7). 
Then steering the flexible zip-tie in a semi-spatial manner, where the zip-tie deformation is carried out in a~planar manner without  torsion (Fig. 8). In the semi-spatial case, the  base frame orientation matrix at $s \!=\! 0$ is $R_x(\phi_x) \!\cdot\! R_Y(\phi_y) \!\cdot\! R(\phi_0)$, and
the distal endpoint frame orientation matrix at $s \!=\! L$ is $R_x(\phi_x) \!\cdot\! R_y(\phi_y) \!\cdot\! R(\phi_0 \!+\! \phi(\mbox{\small $L$}))$.\\
\indent Fig.~7 shows the planar steering of the flexible zip-tie between two obstacles (Fig. 7(a)-(d)), with snapshots of the steering path computation (Fig. 7(e)-(h)). Using weighted $\mbox{\small $A$}^*$ described in Section~IV, the steering path was computed in the flexible cable {\small 6-D} configuration space in $4.5$ seconds. However, one should keep in mind that {\small RRT}  algorithms perform equally well in high dimensional configuration spaces. Hence, we also implemented the bi-directional {\small CBiRRT}  algorithm~\cite{berenson09} on the planar steering problem. When computed over $100$ runs for the specified endpoint states, the  {\small CBiRRT}  algorithm mean execution time was $1.58$ seconds with standard deviation of $0.87$ seconds.\\
\indent Fig.~8 shows the semi-spatial steering of the flexible zip-tie between two cylindrical obstacles (Fig. 8(a)-(d)), with snapshots of the steering path computation (Fig. 8(e)-(h)). The steering path was computed in the flexible~cable~\mbox{\small 9-D} configuration space in $8.6$ seconds. Again,  the 
{\small CBiRRT}  algorithm performed better than weighted $\mbox{\small $A$}^*$.  When computed over $100$ runs for the specified endpoint states, the {\small CBiRRT}  algorithm mean execution time was only $2.31$ seconds with standard deviation of $5.01$ seconds.
Video clips of the experiments are attached to the paper submission and are also available in~\cite{levin&grinberg}.


\begin{figure}
\vspace{.1in}
\centerline{\includegraphics[width=0.5\textwidth]{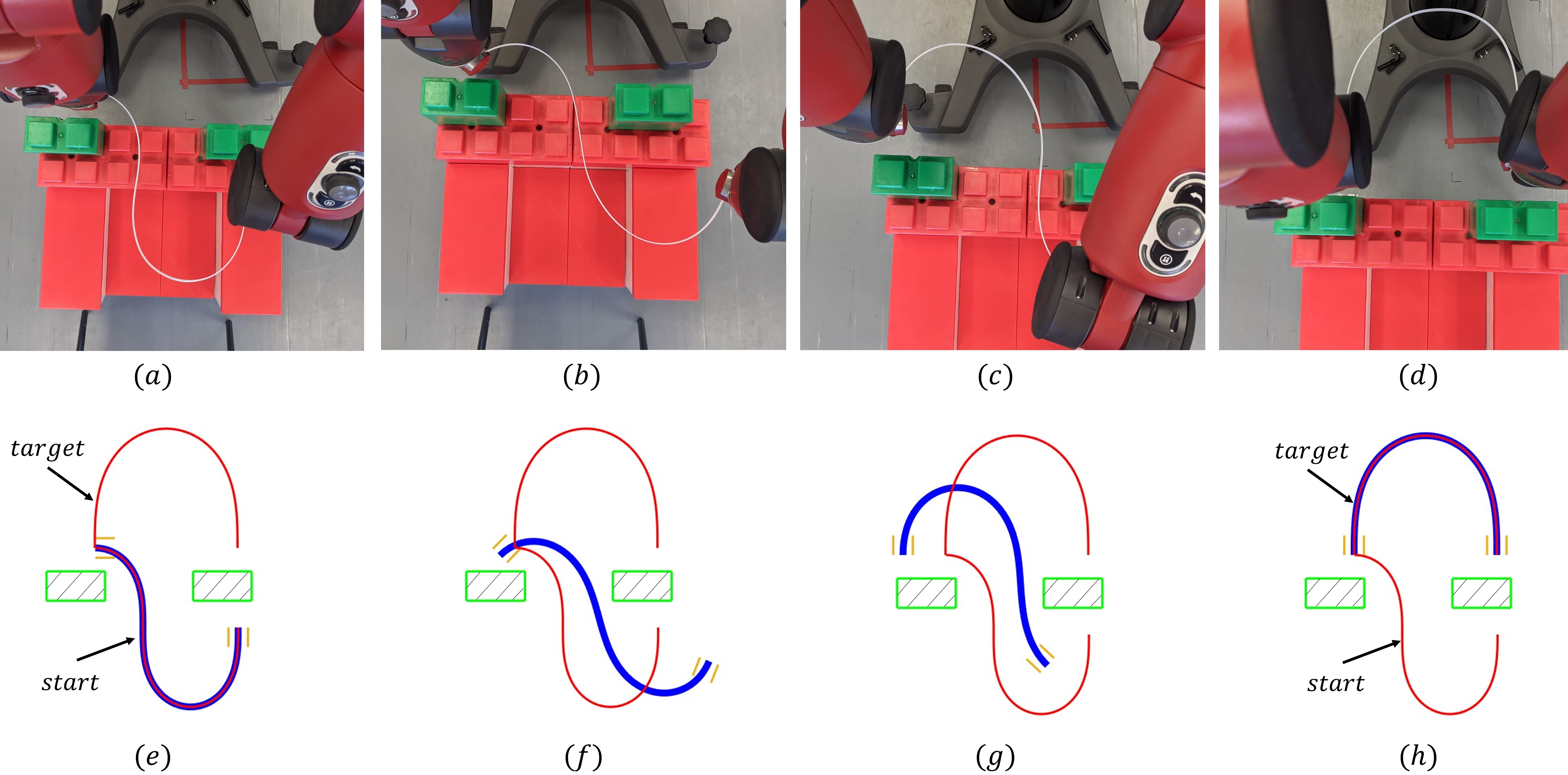}}
\vspace{-.06in}
\caption{Flexible cable steering in a~planar environment between two obstacles. (a)-(d) Baxter robot snapshots. (e)-(h) Matlab simulation of snapshots: red cable is start and target positions, blue cable is current position, green boxes are the obstacles. Note the good fit between the physical zip-tie (upper row) and the elastica prediction shapes (lower row).}
\vspace{-.22in}
\label{fig 8}
\end{figure}



\begin{figure}
\vspace{.1in}
\centerline{\includegraphics[width=0.45\textwidth]{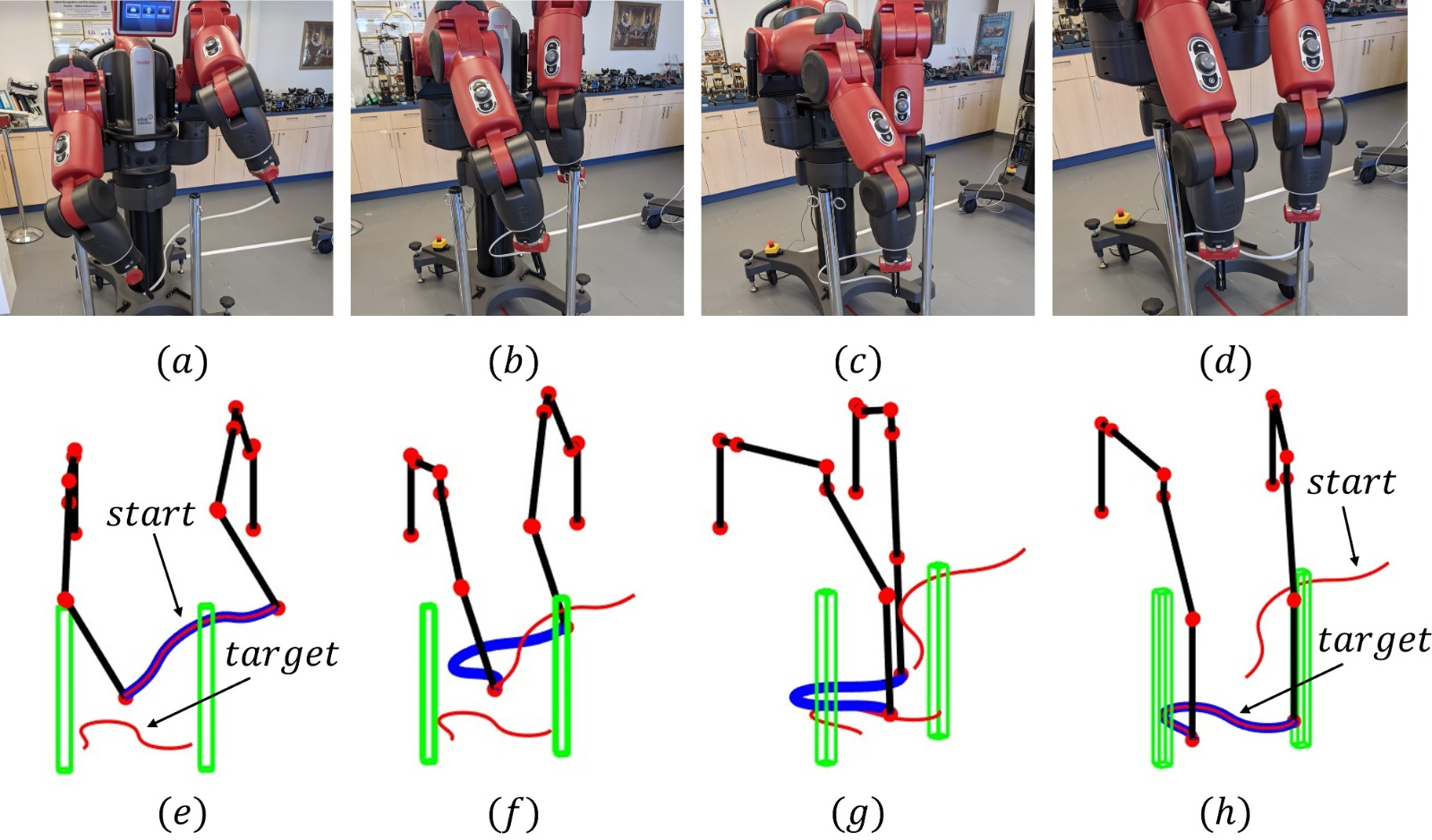}}
\vspace{-.075in}
\caption{Flexible cable semi-spatial steering between two cylindrical obstacles. (a)-(d) Baxter robot snapshots. (e)-(h) Matlab simulation of snapshots: red cable is start and target positions,  blue cable is current position, black lines with red points are the Baxter hands, green cylinders are the obstacles (see video clips attached to submission).}
\label{fig 9}
\end{figure}

\vspace{-.02in}
\section*{\textbf{VI. Conclusion and Future Work}}
\vspace{-.1in}

\noindent The paper developed a steering scheme for flexible cables held by two robot hands that control the cable endpoints position and tangents. 
The flexible cable equilibrium shapes are computed using Euler's elastica solutions. 
The flexible cable configuration space in planar environments consists of the cable base frame position and orientation 
and the elastica parameters $\mathrm{k}$, $s_0$ and $\Tilde{L}$, forming a {\small 6-D} configuration space. The elastica parameters that ensure stability and non self-intersection define the allowed configuration space, $\mathcal{C}_{free}$. To allow flexible cable steering in the presence of obstacles within $\mathcal{C}_{free}$, an~efficient technique that checks collision of the flexible cable convex arcs against obstacles was described. These tools were incorporated into a flexible cable steering scheme that was implemented and demonstrated with experiments. First in planar environments then in three-dimensions where the flexible cable maintained planar formations while moving the flexible cable base-frame position and orientation freely in a {\small 9-D} configuration space.

Future research will extend this work in several ways. First, the range of the elliptic modulus parameter will be extended from $0\leq \mathrm{k} \!<\! \mathrm{k}_{max}$ to any $\mathrm{k}$ in the range $[0,1)$ without self-intersection. This extension will use an~implicit expression that defines the self-intersection points and depends on the elastica parameters.
The second extension will consider flexible cable steering by two robot hands under gravity in two-dimensions. Under gravity, the flexible cable shape minimizes the combined elastic and gravitational potential energies.
The third extension will consider flexible cable steering by a {\em single} robotic hand while the flexible cable distal endpoint is held against a~wall or a~table. Under such single-hand steering, zero moment constraint holds at the contacting tip or segment, captured by the co-state condition $\lambda_\phi(L) \!=\! 0$. Finally, an important practical extension would be to adapt the flexible cable steering method to close-loop steering control. At each step the global planner can compute {\small $N$} forward steps, a single steering step executed,
then the true cable shape measured and used to re-calibrate the global planner for the next {\small  $N$} forward steps.

\vspace{-.04in}
\section*{\textbf{ APPENDIX A -- Proof of Lemma 1}}
\vspace{-.04in}

\noindent This appendix explains why full-period elastica shapes are stable for all $\mathrm{k}\in[0,\mathrm{k}_c]$. The appendix starts with the following auxiliary lemma.
\vspace{.2in}

\noindent\textbf{Auxiliary Lemma [stable equal endpoint tangents shapes]:} {\em When the flexible cable is held with equal endpoint tangents, the flexible cable is 
\vspace{-.04in}
\begin{enumerate}
    \item possibly stable if the cable shape contains one or two inflection point. 
    \item unstable if the cable shape contains three or more inflection points.
\vspace{-.04in}
\end{enumerate}}

\noindent A proof of the auxiliary lemma appears in [33][theorem 1 and [33][Corollary 4.1]. Using the auxiliary lemma, three possibly stable shapes with equal endpoint tangents exist: full period elastica with  $\tilde{L} \!=\! L$, $s_0 \!\in\! [0, L]$ and $\mathrm{k} \!\in\! [0, 1)$; and less than full period shapes with parameters $\tilde{L} \!>\! L$, $\mathrm{k} \!\in\! [0, 1)$ and either $s_0 \!=\! \tilde{L}/4 + (\tilde{L} - L)/2$ or $s_0 \!=\! 3\tilde{L}/4 + (\tilde{L} - L)/2$. The proof of Lemma 1 from Section~III follows.


\indent \textbf{Proof Lemma 1}:
Using numerical means shown in Fig.~9, we show that for all the equal endpoint tangent shapes that are possibly stable according to the auxiliary Lemma 
with base frame $S(0)=(0,0,0)$, flexible cable shapes in the domain $s_0 \!\in\! [0, L], \mathrm{k}\in [0,\mathrm{k}_c]$, and $L=\Tilde{L}$ form a global minimum of the elastic energy. 
Consider the flexible cable total elastic energy given by
\vspace{-.09in}
\[ 
    J= \mbox{\small $\frac{1}{2}EI\cdot\int_0^L \kappa^2(s)ds=2EI\cdot\sqrt{\lambda}P  $} 
\vspace{-.07in}
\]
\noindent where $\mbox{\small $P$} \!=\! \epsilon(\mbox{\small $\sqrt{\lambda}$} (s_0 \!+\! L),\mathrm{k}) \!-\! \epsilon(\mbox{\small $\sqrt{\lambda}$} s_0,\mathrm{k}) \!-\! \sqrt{\lambda}(1 \!-\!\mathrm{k}^2)\mbox{\small $L$}$
and $\epsilon(f(s),\mathrm{k})=E(\mathrm{am}(f(s),\mathrm{k}),\mathrm{k})$.
Based on Fig.~9(a)-(b), for all full-period shapes with $\mathrm{k}>\mathrm{k}_c$ (green points), there exists another full-period shape with $0 \!<\! \mathrm{k} \!<\! \mathrm{k}_c$ (blue points) or less than full-period shape (red points) having the same endpoint constraint that have \textit{a lower} total elastic energy $J$. Additionally, there is {\em no overlap} between the full-period shapes with $\mathrm{k}\!<\! \mathrm{k}_c$ and the shapes with less than full-period. Therefore, the full-period shapes with $\mathrm{k}<\mathrm{k}_c$ are global minima of $J$ and thus form locally stable shapes. It can be concluded that full period shapes with elastica parameters $s_0 \!\in\! [0, \mbox{\small $L$}], \mathrm{k} \!\in\! [0,\mathrm{k}_c]$, and $\mbox{\small $L$} \!=\! \mbox{\small $\Tilde{L}$}$ do {\em not} contain conjugate points along the cable shape.  \epf


\begin{figure}[H]
\vspace{-.08in}
\centerline{\includegraphics[scale=0.3]{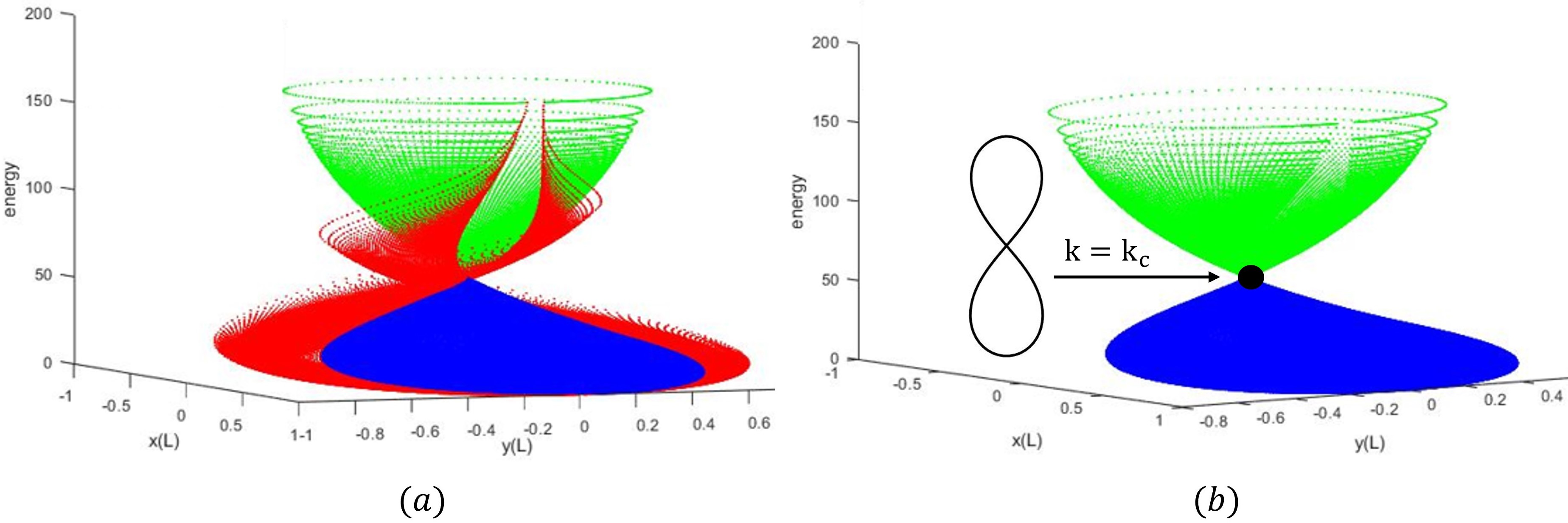}}
\vspace{-.08in}
\caption{Flexible cable total elastic energy with equal endpoint tangents 
plotted above the $(x(L),y(L))$ plane. (a) Blue points are full-period shapes with $0 \leq \mathrm{k}<\mathrm{k}_c$. Red points are stable equal endpoint tangent shapes with less than a full-period, green points are full-period shapes with $\mathrm{k}_c \leq \mathrm{k}\leq 1$. (b) Blue points are full-period shapes with $0 \leq \mathrm{k} < \mathrm{k}_c$. Green points are full-period shapes with $\mathrm{k}_c \leq \mathrm{k}\leq 1$. The energy-level hourglass surface shows the global minimality of the blue points. Black point between the blue and green points represents the full-period \textit{figure eight} shape with $\mathrm{k}=\mathrm{k}_c$. All blue points represent locally stable full-period shapes.}
\label{fig 9}
\vspace{-.12in}
\end{figure}

\vspace{-.13in}
\section*{\textbf{APPENDIX~B -- Neglecting~Gravity During Semi-Spatial Steering}}
\vspace{-.1in}

\noindent This appendix describes a criterion for neglecting the relative effect of gravity during
{\small 3-D} flexible cable steering. Consider the three energy functionals
\begin{align*}
J_T(S, u) &= \int_0^L \big( \frac{1}{2} EI u^2 + \rho g z(s) \big) ds \\
J_E(S, u) &= \int_0^L \frac{1}{2} EI u^2 ds \\
J_G(S, u) &= \int_0^L \rho g z(s) ds
\vspace{-.2in}
\end{align*}  
\noindent where $J_T(S,u)$ represents the flexible cable total elastic and gravity energy with $\rho$ being the linear mass density and $g$ the gravitational constant. Here $z(s)$ is the flexible cable height as a function of its length parameter~$s$. The functional $J_E(S,u)$ represents only the elastic energy stored in the flexible cable, while $J_G(S,u)$ represents only the gravitational energy stored in the flexible cable.

Using the state variables $S(s)=(x(s),y(s),\phi(s))$, the system $\frac{d}{ds}S(s)$ is still described by Eq. (1) for all three cases. Analytic solutions exist for the flexible cable shapes: $(S_E, u_E)$ the solution that minimizes the functional $J_E$ and $(S_G, u_G)$ the solution that minimizes the functional $J_G$.
Now, the flexible cable total energy can be written as 
\vspace{-.08in}
\[
J_T(S, u) = 
\mbox{\small $
J_E(S, u) \!\cdot\! \left( 1 + \frac{J_G(S, u)}{J_E(S, u)} \right) $} 
\vspace{-.09in}
\]
\noindent When $\frac{J_G(S, u)}{J_E(S, u)} \!\ll\! 1$, the flexible cable total~energy~$J_T$ is dominated by the cable elastic energy $J_E$.
Furthermore, it holds that $J_G(S_G, u_G) \!\leq\! J_G(S, u)$ and $J_E(S_E, u_E) \!\leq\! J_E(S, u)$ for all $S$ and $u$. The ratio $J_G(S_E, u_E)/J_E(S_E, u_E)$ thus forms a {\em conservative estimate}. When the ratio $\tfrac{J_G(S_E, u_E)}{J_E(S_E, u_E)} \!\ll\! 1$, the effect of gravity can be neglected as illustrated in the following example.

{\bf Example:} The flexible zip-tie used in the experiments of Section~V has a rectangular cross section of dimensions $1 \!\times\! 9$~$\mathrm{mm}$. Its linear mass density is $\rho \!=\! 0.013$~$\mathrm{Kg/m}$  and its bending stiffness is $EI \!=\! 0.0027$~$\mathrm{N m^2}$. As a representative example consider the elastica parameters
$\mathrm{k} \!=\! 0.87367$, $s_0 \!=\! 0.13 \Tilde{L}$ and $\Tilde{L} \!=\! 1.1 L$.  For these parameters, $J_G(S_E, u_E) / J_E(S_E, u_E) \!=\! 0.0047$ when the zip-tie length is $L \!=\! 50$~$\mathrm{cm}$  and  
$J_G(S_E, u_E) / J_E(S_E, u_E) \!=\! 0.0375$ when the zip-tie length increases to $L \!=\! 100$~$\mathrm{cm}$. $\hfill \circ$


\vspace{-.07in}

\bibliographystyle{IEEEtran}

\small{
\begin{bibliography}{books,elonbib,elon}
\end{bibliography}
}
\end{document}